\documentclass[twocolumn,english,natbib=False,sigconf,letterpaper,9pt]{acmart}
\usepackage[sort&compress]{natbib}
\acmConference[FAT*'19]{ACM Conference on Fairness, Accountability, and Transparency}{January 2019}{Atlanta, Georgia USA} \acmYear{2019} \copyrightyear{2019}
\setcopyright{rightsretained}
\acmBooktitle{FAT* '19: Conference on Fairness, Accountability, and Transparency (FAT* '19), January 29--31, 2019, Atlanta, GA, USA}
\acmPrice{15.00}
\acmDOI{10.1145/3287560.3287574}
\acmISBN{978-1-4503-6125-5/19/01}



\begin{document}
\title{Explaining Explanations in AI
}

\author{Brent Mittelstadt}     
\email{brent.mittelstadt@oii.ox.ac.uk}
\affiliation{\institution{University of Oxford}}  
\affiliation{\institution{The Alan Turing Institute}} 

\author{Chris  Russell}    
\email{crussell@turing.ac.uk}
\affiliation{\institution{University of Surrey}}
\affiliation{\institution{The Alan Turing Institute}}  

\author{Sandra Wachter}      
\email{sandra.wachter@oii.ox.ac.uk}
\affiliation{\institution{University of Oxford}} 
\affiliation{\institution{The Alan Turing Institute}}      
     
\renewcommand{\shortauthors}{Mittelstadt, Russell, Wachter}

\begin{abstract}
Recent work on interpretability in machine learning and AI has focused on the building of simplified models that approximate the true criteria used to make decisions. These models are a useful pedagogical device for teaching trained professionals how to predict what decisions will be made by the complex system, and most importantly how the system might break. However, when considering any such model it's important to remember Box's maxim that "All models are wrong but some are useful." We focus on the distinction between these models and explanations in philosophy and sociology. These models can be understood as a "do it yourself kit" for explanations, allowing a practitioner to directly answer "what if questions" or generate contrastive explanations without external assistance. Although a valuable ability, giving these models as explanations appears more difficult than necessary, and other forms of explanation may not have the same trade-offs.  We contrast the different schools of thought on what makes an explanation, and suggest that machine learning might benefit from viewing the problem more broadly.
\end{abstract}\maketitle

\section{Introduction}
As we deploy automated decision-making systems in the real world, questions of accountability become increasingly important.\footnote{This work was supported by The Alan Turing Institute, EPSRC grant EP/N510129/1, and the British Academy, grant PF170151.} For system builders questions such as ``Is the system working as intended?'', ``Do the decisions being made seem sensible?'' or ``Are we conforming to equality regulation and legislation?'' are important, while a subject of the decision-making algorithm may be more concerned with topics such as ``Am I being treated fairly?'' or ``What could I do differently to get a favourable outcome next time?''

These issues are not unique to computerised decision-making systems, but with the growth of machine learning based systems they have become even more important \citep{pasquale_black_2015,bodo_tackling_2017,kroll_accountable_2016,veale_clarity_2018,nissenbaum_accountability_1996,selbst_intuitive_2018,olhede_growing_2018}. What distinguishes machine learning is its use of arbitrary black-box functions to make decisions. These black-box functions may be extremely complex and have an internal state composed of millions of interdependent values. As such, the functions used to make decisions may well be too complex for humans to comprehend; and it may not be possible to completely understand the full decision-making criteria or rationale. 

Under these constraints, it becomes an open question as to what forms of explanation are even possible that can answer the earlier questions. As such, one of the most striking aspects of research into explainable AI (xAI) is how many different people, be they lawyers, regulators, machine learning specialists, philosophers, or futurologists, are all prepared to agree on the importance of explainable AI. However, very few stop to check what they are agreeing to, and to find out what explainable AI means to other people involved in the discussion \citep{Lipton2016Mythos}. 

This gap of expectations is largest between machine learning, which has essentially re-purposed the term ``explanation,'' and the fields of law, cognitive science, philosophy and the social sciences (which we refer to here collectively as the `explanation sciences'), where it has a relatively well-defined technical meaning and a plethora of research on types of explanations, their purposes, and their social and cognitive function.\footnote{The dichotomy between explanations in machine learning and elsewhere is well illustrated by the recent work by \citet{Miller2017Explainable} who found that none of the papers on a curated reading list for xAI made use of work from the social sciences on explanations.} Work on xAI currently occupies only one or two small branches of this diverse research landscape. Specifically, the vast majority of work in xAI produces simplified approximations of complex decision-making functions. We argue that these approximations function more like scientific models than the types of scientific and `everyday' explanations considered in philosophy, cognitive science, and psychology. 

In this paper we examine the extent of this gap between xAI and the `explanation sciences'. We do so by first reviewing methods for producing explanations in xAI, and explain how they are generally more akin to scientific modelling than explanation giving. If this comparison holds, it follows that the majority of currently available methods will, at best, produce locally reliable but globally misleading explanations of model functionality. We then examine research on explanations in philosophy, cognitive science, and the social sciences which suggests that `why-questions' (e.g. `why did the model exhibit that behaviour?') require explanations that are contrastive, selective, and socially interactive. On this basis, we argue that, if xAI is to produce methods that make algorithmic decision-making systems more trustworthy and accountable, the field's attention must shift to the development of interactive methods for post-hoc interpretability that make it easier to contest algorithmic decisions, and facilitate informed dialogue between users, developers, algorithmic systems, and other stakeholders.

\section{A brief primer on explanations in philosophy}
Our interest in the philosophical treatment of explanations is based on observation of the development of the field of xAI, or research addressing interpretability and explainability in machine learning. Our aim is to determine whether xAI is heading in a direction in which explanations can be produced that allow affected parties, regulators, and other non-insiders to understand, discuss, and potentially contest decisions made by black-box algorithmic models. So our primary question is: which types of explanations are currently being produced by xAI? And, are these explanations actually useful to the individuals (or proxies thereof) affected by black-box decisions?

Explanations, and more broadly epistemology, causality, and justification, have been the focus of philosophy for millennia, making a complete overview of the field unfeasible. What follows is a brief review of key distinctions and terminology relevant to our interest in methods for providing explanations of black-box algorithmic models and decisions. This primer broadly follows the structure established in a recent review of the contribution of the social sciences to xAI (see: \citet{Miller2017Explanation}).

Briefly, the xAI community investigates interpretability (or explainability) and ways of providing explanations of algorithmic models and decisions. `Interpretability' refers to the degree of human comprehensibility of a given `black-box' model or decision \citep{Miller2017Explanation, Lisboa2013Interpretability}. Poorly interpretable models ``are opaque in the sense that if one is a recipient of the output of the algorithm (the classification decision), rarely does one have any concrete sense of how or why a particular classification has been arrived at from inputs'' \citep[p.1]{Burrell2016How}. 

In contrast, `explanation' refers to numerous ways of exchanging information about a phenomenon, in this case the functionality of a model or the rationale and criteria for a decision, to different stakeholders \citep{Lipton2016Mythos, Miller2017Explanation}. Explanations of machine learning models and predictions can serve many functions and audiences. Explanations can be necessary to comply with relevant legislation \citep{doshi-velez_accountability_2017}, verify and improve the functionality of a system (i.e. as a type of `debugging'; \citep{Kulesza2015Principles}) and help developers and humans working with a system learn from it \citep{Samek2017Explainable}, and enhance the trust between individuals subject to a decision and the system itself \citep{hildebrandt_challenges_2010,zarsky_transparent_2013,citron_scored_2014}. As these purposes suggest, explanations can be offered to expert developers, professionals working in tandem with a system (e.g. expert labelers of training cases; see \citet{Berendt2017Toward} for an overview of human actors involved in algorithmic decision-making), and to individuals or groups affected by a system's outputs \citep{Weller2017Challenges,mantelero_personal_2016}.

Our interest here is solely in ways of providing explanations. Returning to philosophy, types of explanations can be distinguished according to their completeness, or the degree to which the entire causal chain and necessity of an event can be explained \citep{Ruben2004Explaining}. Often this is expressed as the difference between `scientific' and `everyday' explanations (both of which deal with causes of an event; e.g. \citet{Miller2017Explanation}, or `scientific' (full) and `ordinary' (partial) causal explanations \citep{Ruben2004Explaining}. Miller argues that `everyday explanations' address ``why particular facts (events, properties, decisions, etc.) occurred,'' rather than general scientific relationships \citep[p. 5]{Miller2017Explanation}.

While these distinctions hide significant nuance, they matter insofar as they constrain the scope of our focus in discussing explanations in AI. In recent calls for explanations in AI, and in work on interpretability in machine learning more broadly, explanations are requested in connection to a particular entity, be it a specific decision, event, trained model, or application. The explanations requested are thus not full scientific explanations, as they need not appeal to general relationships or scientific laws, but rather at most to causal relationships between the set of variables in a given model \citep{woodward_explanation_1997}. As such, xAI is effectively calling for everyday explanations either of how a trained model functions in general, or how it behaved in a particular case.

\section{Explainable AI}
Much recent work has been dedicated to rendering machine learning models interpretable or explainable. Two broad aims of work on interpretability have been recognised in the literature: transparency and post-hoc interpretation. Transparency addresses how a model functions internally, whereas post-hoc interpretations concern how the model behaves \citep{Lipton2016Mythos, Lepri2017Fair, Montavon2017Methods}. Transparency can be further specified according to its target. Respectively, a mechanistic understanding of the functioning of the model (simulatability), individual components (decomposability), and the training algorithm (algorithmic transparency) can be sought. Models can be rendered transparent by explanations at a minimum of three levels: ``at the level of the entire model, at the level of individual components (e.g., parameters), and at the level of a particular training algorithm'' \citep{Lepri2017Fair}. A model, its component parts, or its training/learning algorithm can thus be said to be transparent if their functionality can be comprehended in their entirety by a person \citep{Lipton2016Mythos}. 

Post-hoc human interpretable explanations of models and specific decisions do not seek to reveal how a model functions, but rather how it behaved, and why. According to \citet{Lipton2016Mythos}, approaches to post-hoc interpretability include verbal (natural language) explanations (e.g. \citet{McAuley2013Hidden}), visualisations and interactive interfaces (e.g. \citet{Tamagnini2017Interpreting, Simonyan2013Deep}), local explanations or approximations (e.g. \citet{Ribeiro2016Why, Fong2017Interpretable}), and case-based explanations (e.g. \citet{Caruana1999Case-based, Kim2014bayesian}). 

Natural language explanations can consist of ``textual or visual artefacts that provide qualitative understanding of the relationship'' between features of an input (e.g. words in a document) and the model's output (e.g. a classification or prediction; \citep{Ribeiro2016Why}. Visualisation techniques can visually demonstrate the relative influence of features or particular pixels (e.g. in the case of an image classifier), or provide an interface for users to explore textual or visual explanations \citep{Tamagnini2017Interpreting,poulin_visual_2006}. Local explanations seek to explain how a fixed model leads to a particular prediction, either by fitting a simpler, local model around a particular decision \citep{Ribeiro2016Why}, or by perturbing variables to measure how the prediction changes \citep{Simonyan2013Deep, Datta2016Algorithmic, Adler2016Auditing}. Case-based explanation methods \citep{Caruana1999Case-based,Kim2014bayesian} for non-cased based machine learning involve using the trained model as a distance metric to determine which cases in the training data set are most similar to the case or decision to be explained. These training cases can then be shared with parties affected by the decision.

Despite this variety of approaches, a significant amount of the xAI community now pursues methods to retro-fit local or approximate models over more complex algorithms. These simplified models approximate the true criteria used to make decisions (e.g. \citet{Simonyan2013Deep, Ribeiro2016Why, Selvaraju2016Grad-CAM:,Baehrens2010How}). Broadly speaking there are two widely used classes of model \emph{(i)} Linear or Gradient-based approximations that assign a single importance weight to each feature (be it someone's age, or a particular pixel in an image) and \emph{(ii)} Decision tree-based methods that use nested sets of yes/no decisions to approximate classifiers.

These methods can both be applied to create approximations at a global\footnote{Offering a simplified model or a set of simplified models that approximates decisions made for all possible datapoints.} or local\footnote{A simplified model that only approximates decisions made about a few datapoints, typically only a single exemplar.} level. Historically, much work has focused on global approximations of models (e.g. \citet{Sanchez2015Towards, Craven1996Extracting, Martens2007Comprehensible}, including approaches based on clustering \citep{Chen2016Interpretable}, integer programming \citep{Zeng2017Interpretable}, and rule lists \citep{Wang2015Falling}). In contrast, local approximations are accurate representations only of a specific domain or `slice' of a model. A trade-off inherently occurs between the insightfulness of the approximated model, the simplicity of the presented function, and the size of the domain to which is applies and remains valid \citep{Bastani2017Interpretability, Lakkaraju2017Interpretable}. 

No matter the approach taken in xAI, reflexivity is needed to ensure the community actually works towards its normative and practical goals to render models holistically transparent or provide high-quality post-hoc interpretations of model behaviour. Critical questions must be repeatedly asked and answered. For example, will the methods developed make machine learning models more interpretable? More trustworthy to users? More accountable? And to whom will explanations be accessible, comprehensible, and useful? 

Answering such questions requires considering the methods developed in xAI in the context of prior work in fields addressing such normative and social questions. Local and approximation models may in fact resemble existing, well-known approaches to explanations in the 'explanation sciences', which would provide insight into their practical value and limitations for users, developers, and other stakeholders going forward. 

\subsection{Scientific Modelling and Explainable AI}
We believe that the closest analogue for the bulk of methods currently occupying xAI researchers lies in the use of scientific modelling, or the building of approximate models that are not intended to capture the full behaviour of physical systems but rather to provide coarse approximations of how the systems behave. These approximations are useful to experts both for pedagogical purposes and for making reliable predictions of how the system might behave over a restricted domain, but can be misleading when presented as an explanation of how the model functions to a lay user. 

One famous example of this problem is Newtonian physics, which is taught first to schoolchildren and provides a good enough description for much day-to-day engineering, but that famously breaks down as an approximation when high-precision is required at either very large or very small scales, where either general relativity or quantum physics are necessary.\footnote{Famously, GPS satellites are insufficiently accurate unless they account for effects of general relativity.} Both general relativity and quantum physics are also examples of such models. Although extremely accurate in their domains, outside of them they break down, and a unified model of physics that is accurate at all scales is still being sought.

Much scientific theory can be understood as the use and characterisation of such models \citep{1995Theories, Frigg2006Scientific,box1979robustness}. Although any physical system can be understood in terms of the emergent properties of subatomic particles, such descriptions are neither human comprehensible nor computationally feasible. Instead, scientists deal in local approximations that provide accurate descriptions of the phenomena they are interested in, but which may prove inaccurate in a larger domain.

It is in this context that Box's maxim, ``All models are wrong, but some are
useful,'' \citep{box1979robustness} should be understood. Explainable AI generates approximate simple models and calls them `explanations', suggesting reliable knowledge of how a complex model functions. 

When characterising the use of such models in science, \citet{Hesse1965Models} divides the properties of the model into positive analogies, where the properties of the model are known to correspond to properties of the phenomena we are interested in; negative analogies, where the properties of the model do not match the phenomena we are interested in; and neutral analogies, where it is unknown if the properties of the model correspond to the phenomena.

This characterisation captures many of the challenges when offering approximations of models as explanations. It is not enough to simply offer a human interpretable model as an explanation. For an individual to be able to trust such a model as an approximation, they must know over which domain a model is reliable and accurate, where it breaks down, and where its behaviour is uncertain. If the recipient of a local approximation does not understand its limitations, at best it is not comprehensible, and at worst misleading. 

This is not to say that local approximations are without merit, but rather that they can only reliably have explanatory power if their limitations are clearly documented and understood by recipients. For domain experts with in-depth knowledge of when and where approximations break down, or for technicians that have a clearly defined and tested remit for where the approximation can be used, they can be extremely useful. However, at the moment xAI generally avoids the challenges of testing and validating approximation models, or fully characterising their domain. If these elements are well understood by the individual, models can offer more information than an explanation of a single decision or event. Over the domain for which the model accurately maps onto the phenomena we are interested in, it can be used to answer `what if' questions, for example ``What would the outcome be if the data looked like this instead?'' and to search for contrastive explanations, for example ``How could I alter the data to get outcome X?''

However, local models may also provide false assurances. As suggested above, local approximations are often misleading or inaccurate outside of their domain, and provide little insight into how a function response and outcomes vary with changes to the inputs. By definition local explanations hold only for a specific decision; what is explained is not how the model functions as a whole, but rather one segment of the model relevant to the prediction at hand. Thus, while helpful to explain the weights and relationships between variables in a small segment of a model (relevant to a particular case or decision), the explanations do not provide evidence of the trustworthiness or acceptability of the model overall. 

\subsection{Linear approximations}
To provide further support for the analogue between xAI and scientific modelling, we turn now to variants of linear approximation. Many of the issues and design decisions made by researchers in the field point directly towards the issues with modelling previously, particularly the three-way trade-off between the simplicity of the approximated model, the size of the domain it describes, and the accuracy of this description.

\subsubsection{Linear Models in Continuous Spaces}
Linear models are designed to give a single measure of the importance of each variable to the classifier they are approximating. In some cases (e.g. \citet{Ribeiro2016Why, Lundberg2017Unified}, these weights can be directly interpreted as the sensitivity or a number that tells you how much the classifier response will vary as a particular feature changes. In other words, if a particular variable has a weight of  associated with it, altering the variable by small amount  will cause the classifier to vary by approximately . Regardless of whether linear models are intended to be a local approximation of a classifier or a simple importance measure, they suffer from two distinct issues. The first is a problem of curvature, namely that the sensitivity of a classifier to a change in a particular variable may vary with the amount the variable changes. The second issue is one of dependencies between variables and how to capture the relationships between them. 

Both of these issues can be illustrated by an example taken from \citet{Miller2017Explanation}. Suppose, I believe that a particular creature is a bee because I have been told it has 6 legs and four wings.  I am unlikely to change my mind if I am told it actually has three wings. The most reasonable explanation for a three winged insect is that it was originally a four winged insect that lost a wing, but if I'm told that it has two wings I may well believe that it is a fly. This is an example of the sensitivity varying with the size of the change made. On the other hand, I will only be comfortable believing it is a spider if it has both eight legs and no wings. Should I then say that my belief depends wholly on the number of legs it has; wholly on the number of wings; or weakly on both? As linear models do not capture the interdependencies between variables, they cannot accurately characterise these relationships. In all such cases, when a local model is fitted to the classifier response, the choice of domain or variable values governs how well the approximation performs.

\subsubsection{Gradient Sensitivity verses Binarization}
There are two main schools of thought regarding the problems of varying sensitivity and scale. The first notes that the sensitivity of a classifier is only well defined when the size of  in the previous section tends towards zero. In this case, the sensitivity is equivalent to the gradient. Although well defined, this essentially means that the model is fitted to a domain of size 0 and may exhibit large amounts of instability while offering limited predictive power. Two prominent approaches are described by \citet{Simonyan2013Deep} and \citet{Baehrens2010How}. 

The other option, and most influentially proposed by \citet{Ribeiro2016Why} in their Locally Interpretable Model-Agnostic Explanations (LIME) approach, is to binarize the problem. Rather than trying to fit a linear classifier to a large range of values, the authors consider a binary problem, where for each feature they attempt to switch it on and off, allowing them to answer the question ``What is the contribution of feature $f$ to the classifier response, given the data it currently sees?'' This leaves open the question of ``the contribution compared to what?'' For unstructured data, such as a count of how many times particular words occur in a document, it makes sense to compare against a baseline created by setting the count to 0. For structured data this is more problematic. For example, how can we evaluate the importance of someone's salary to a loan decision, if the classifier can only evaluate people with valid salaries? The answer is to compare it against a different valid salary, but it is unclear how this valid salary should be chosen.

This issue is even more apparent when creating local approximations of computer vision algorithms, as individual pixels cannot be removed from an image, but only set to different values. Several options have been proposed. LIME appeared to set regions of the image to an unspecified homogeneous value \citep{Ribeiro2016Why}. Deep Taylor Decomposition \citep{Montavon2017Methods} suggests blurring the image, an operation which preserves colour information but removes texture. DeepLift uses a user-specified value \citep{Shrikumar2016Not, Shrikumar2017Learning}, while layerwise relevance propagation sets internal network values to 0 \citep{Montavon2017Methods}. Each of these choices is an implicit restriction of the domain over which the model is fitted and carries different implications for the kinds of model found, and can substantially alter the importance given to features. For example, contrasting current image values against a particular colour, such as grey, makes it appear that grey pixels have no effect, while contrasting an image against its blurred version makes it appear as if only high-frequency texture cues are important and that the classifier does not use colour information.

\subsubsection{Linear models in high-dimensional spaces}
\begin{figure*}
  \includegraphics[width=2\columnwidth]{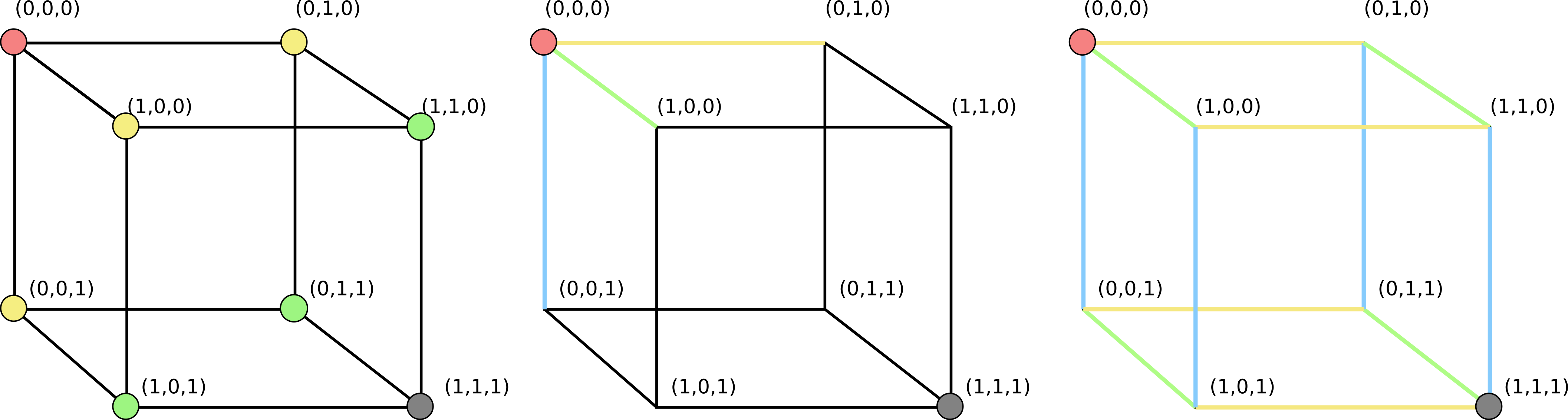}
\caption{ An illustration of the different weighting scheme used in fitting
  linear models. LIME (leftmost) weights all examples differently based on how
  far they are from the original data point (illustrated by different coloured
  vertices), while DeepLift (centre) only fits the linear weights to the
  individual edges closest to the original data point (coloured pink). SHAPE
  (right) fits weights by averaging over all edges formed by flipping a single
  variable from off to on (each group averaged together is indicated by a single
  colour). Each of these different approaches equates to a different assumption
  as to which samples are most important, and none of them can be said \textit{a priori} to be
  better than any of the others.\label{fig1}}
\end{figure*}

Having made a choice of a binarization as discussed in the previous section, a question then remains as to which of these values to approximate with a linear model. Even after restricting a function defined over a continuous high-dimensional range to a set of binary variables, this gives a (hyper-)cube of  possible values while a linear function can only uniquely specify  values, with all other values being a linear approximation of that (see Figure \ref{fig1} for an illustration). This raises the question of which of these values are important to approximate. For example, should we only pay attention to contrasting solutions closest to the original solution as suggested by DeepLift (illustrated by Figure \ref{fig1} centre; \citep{Shrikumar2016Not, Shrikumar2017Learning}); uniformly weight over all possible values, suggested by SHAPE \citep{Lundberg2017Unified} (Figure \ref{fig1} right); or pay as much attention to values close to the data point as to those close to the alternate solution (e.g. a solid grey image); or a weighted mixture of the two approaches (e.g. Figure \ref{fig1} left LIME; \citep{Ribeiro2016Why})?

\subsection{Exploring alternatives to scientific modelling}
Local approximations thus face difficulties with generalizability, arbitrariness in choice of domain, and the potential to mislead recipients unless the domain and epistemic limitations of the approximation are known. Given these difficulties, other methods of producing explanations may be preferable from the perspective of the user or individual affected by a black-box system. The utility of local approximations is dependent upon the knowledge of the recipient regarding the approximations limitations, including conditions under which it will break down and provide a misleading explanation of the decision-making model. Local approximations can be useful as a type of `explanation kit' or causal chain that allows expert users to explore slices of a model for prototyping or debugging \citep[p.17]{Miller2017Explanation}\citep{poulin_visual_2006}. However, their ultimate utility and reliability for non-experts, including individuals subject to decisions made by the system, is highly questionable. 

This finding raises the question: might other methods for generating explanations perform better, or at least offer different benefits, than local approximations? Can other methods provide more reliable or personally relevant information for non-experts to enhance accountability and trustworthiness in algorithmic systems?

\section{Contrastive Explanations}
Thus far we have discussed the philosophical and practical purposes of scientific models, which can be understood as partial causal scientific explanations that assist in comprehending a piece of  the functionality of a phenomenon \citep{Ruben2004Explaining}. Given the difficulties faced by local approximations, it is worth examining prior work in the 'explanation sciences' to identify potential alternative approaches to generate reliable and practically useful post-hoc interpretations for parties affected by an algorithmic decisions. If our goal is to produce explanations that are comprehensible and useful to expert as well as non-expert stakeholders, it is sensible to examine theoretical as well as empirical work describing how humans give and receive explanations \citep[p.3-4]{Miller2017Explanation}. 

In recent decades, work in the philosophy of science and epistemology has paid increasing attention to theories of contrastive explanations and counterfactual causality (e.g. \citet{Ruben2004Explaining, Lewis1973Counterfactuals,Woodward2003Scientific,kment_counterfactuals_2006}). In short, contrastive theories argue that causal explanations inevitably involve appeal to a counterfactual case, be it a cause or event, which did not occur. A canonical example is provided by Lipton \citet{Lipton1990Contrastive}: ``To explain why P rather than Q, we must cite a causal difference between P and not-Q, consisting of a cause of P and the absence of a corresponding event in the history of not-Q'' . Some authors go so far as to claim all questions about causality are inherently contrastive \citep{Ruben2004Explaining, Lewis1973Counterfactuals}.

Contrastive theories of explanation are of course not without criticism. \citet{Ruben2004Explaining}, for example, has suggested that, even if causal explanations are inevitably contrastive in nature (which he doubts), this characteristic can be dealt with by traditional theories of explanation, rendering the `contrastive turn' interesting but ultimately unnecessary. While the utility of contrastive theories remains debated, the fact that contrastive explanations address a particular event or case and are thus simpler to generate than complete or global explanations of model functionality suggest they worth further consideration in xAI \citep{Lipton1990Contrastive}.

A recent review by \citet{Miller2017Explanation} suggests substantial empirical support exists for the practical utility of `everyday' contrastive explanations. Miller reviewed articles and empirical studies from ``philosophy, psychology, and cognitive science of how people define, select, evaluate, and present explanations'' \citep[p.1]{Miller2017Explanation}. His analysis highlighted three primary characteristics of explanations as they are used, selected, evaluated, and shared by individuals: 

\subsection{Human explanations are contrastive}
`Everyday explanations' are ``sought in response to particular counterfactual cases...That is, people do not ask why event P happened, but rather why event P happened instead of some event Q'' \citep[p.5]{Miller2017Explanation}.\footnote{It is worth noting that counterfactual cases in contrastive, everyday explanations of algorithmic decisions are not equivalent to counterfactuals used to assess causality \citep[p.13]{Miller2017Explanation}\citep{Hilton1986Knowledge-based,woodward_explanation_1997}, in the sense that the range of possible alternatives is necessarily bounded by the limitations of the model in question and the features available to it. This invariance in the model allows for reliable contrastive or counterfactual explanations to be computed \citep{woodward_explanation_1997}.} The preference for contrastive explanations is not due merely to the cognitive complexity of non-contrastive explanations, for example the number of links in a causal chain. Rather, the reviewed empirical evidence indicates that humans psychologically prefer contrastive explanations \citep[p.18]{Miller2017Explanation}\citep{Rehder2003causal-model, Rehder2006When}. 

The perceived abnormality of an event influences requests for contrastive explanations which address why a normal or expected event did not occur \citep{Hilton1986Knowledge-based, Samland2014Do,McClure2003Role}. `Normal' behaviour has empirically been shown to be judged as ``more explainable than abnormal behaviour,'' with perceived abnormality playing an important role in explanation selection \citep[p.41]{Miller2017Explanation}.  \citet{Gregor1999Explanations} support the importance of abnormality, suggesting that users request explanations when an anomaly or abnormal event is detected. \citet{lim_assessing_2009} similarly note a positive relationship between the perceived "inappropriateness" of application behaviour and user requests for contrastive explanations. Violation of ethical and social norms can likewise set an event apart as abnormal \citep{Hilton1996Mental}. Explanations addressing why an alternative, expected event did not occur have historically been addressed in the experts systems literature, seen for instance in the discussion of "Why Not" explanations by \citet{lim_assessing_2009}.

\subsection{Human explanations are selective}
Full or scientific explanations are rarely if ever realised in practice. Absent general laws or the complete causal chain leading to an event, multiple explanations are typically possible that attribute different causes. A given cause of set of causes may be incomplete insofar as they are not the sole cause of the event, but nonetheless convey useful information to the recipient in a given context or for a given purpose \citep{Ylikoski2013Causal}. As Miller argues, ``Explanations are selected -- people rarely, if ever, expect an explanation that consists of an actual and complete cause of an event. Humans are adept at selecting one or two causes from a sometimes infinite number of causes to be the explanation'' \citep[p.5]{Miller2017Explanation}. Further, to be informative, explanations should not be entirely reducible to presuppositions, or beliefs that the recipient of the explanation already holds \citep{Hesslow1988problem}. They should further be relevant to the question  asked by the recipient (epistemic relevance; \citep{Miller2017Explanation}, or what \citet{Slugoski1993Attribution} describe as the recipient's context.

When an explanation giver ('explainer') selects an explanation for an event, possible or actual causes can be `backgrounded' or 'discounted', meaning they are disregarded on the basis of contextual information that renders them irrelevant to the purposes of the explainer or recipient of an explanation (the 'explainee'). This type of selection is essential to reduce long causal chains to a cognitively manageable size \citep{Hilton1996Mental}. In xAI, selection often takes the form of key features or evidence being emphasised in explanation interfaces based upon their relative weight or influence on a given prediction or output \citep{poulin_visual_2006,biran_justification_2014}. As the observations above suggest, the relevance of features (and explanations addressing them) would be based not only on 'statistical weight', but also the explainee's subjective interests and expectations.

\subsection{Human explanations are social}
Explanations are social, insofar as they involve an interaction between one or more explainers and explainees. Interactive transfer of knowledge is required in which information is tailored according to the recipient's beliefs and comprehensional capacities \citep[p.5]{Miller2017Explanation}. Explanations can be conceived as involving one or more explainers and explainees engaging in information transfer through dialogue, visual representation, or other means \citep{Hilton1990Conversational}, often to correct information or knowledge assymetry \citep{lim_assessing_2009}. In the case of machine learning models, it is perhaps most useful to always treat explanation generation as an interactive process, initially involving a mix of human and automated actors, at a minimum an inquirer (e.g. a developer, user) and the model or system \citep{Kayande2009How, Martens2013Explaining}. Further, explanations are iterative, insofar as they must be selected and evaluated on the basis of shared presuppositions and beliefs. Relevance is key, and iteration may be required to communicate effectively or clarify points of confusion on the path towards a mutually understood explanation.

Together, these characteristics of everyday explanations reveal that they ``are not just the presentation of causes (causal attribution). While an event may have many causes, often the explainee cares only about a small subset (relevant to the contrast case), the explainer selects a subset of this subset (based on several different criteria), and explainer and explainee may interact and argue about this explanation'' \citep[p.6]{Miller2017Explanation}.

\subsection{Contrastive explanations in xAI}
Contrastive methods of generating explanations are responsive to these three characteristics of explanations emphasised in the 'explanation sciences'. Two approaches for directly computing contrastive explanations are described by \citet{Martens2013Explaining} and \citet{Wachter2018Counterfactual}.\footnote{These methods resemble the aforementioned "Why Not" explanations described by \citet{lim_assessing_2009}, and are related to work on adversarial perturbations (e.g. \citep{goodfellow_explaining_2014,dube_high_2018}} Such post-hoc methods avoid many of the difficulties faced by model-based explanations. Rather than explicitly generating a model that approximates functional values over a restrictive domain, and relying on the user to interpret this, contrastive explanations directly offer an alternative data point: ``If your data had looked like this, you would have been given this classification score instead.'' These alternative data points can be computed exactly. As such, many of the challenges facing 'modelling' approaches to generating explanations, such as the quality of the approximation or the limits of a chosen domain, do not arise to a comparable degree. 

However, as contrastive methods only return a single data point, a more pressing concern is the relevance of the output. If this data point does not directly correspond to a factoid of interest to the user, it cannot be used to deduce relevant conclusions, for example regarding the justifiability of a decision. Similar issues arise in the fitting of models: if the domain the model is fitted to does not capture examples relevant to the intended audience, then it is unlikely to be useful.

The first approach described by \citet{Martens2013Explaining} is explicitly designed for use on discrete data, and focuses on the particular problem of which words need to be removed from a website in order for it to be no longer be classified as an adult (i.e. pornographic) website. In contrast, \citet{Wachter2018Counterfactual} propose a method, 'counterfactual explanations', designed to work on primarily continuous data. They illustrate their approach on the problem of law school admissions and risk factors likely to increase a patient's chance of developing diabetes.

Finding such counterfactuals explanations can be described as a search or optimisation problem. Such approaches seek a similar counterfactual that is both close to the original datapoint and likely to occur in the real world \citep{kment_counterfactuals_2006}. In the case of \citet{Wachter2018Counterfactual} this was formulated as a Lagrangian style constrained optimisation: \begin{equation}
\arg\min_{\bf c} \max_\lambda \lambda(f({\bf c})-T)^2 + d({\bf c},{\bf x})
\end{equation}

Where $\bf x$ is the original data point, and $\bf c$ the counterfactual.
$f({\bf c})$ is the classification response from the black-box function $f$, which is constrained to take target value $T$. $d(\cdot,\cdot)$ is a distance function that ensures that the counterfactual is a relevant, and a human comprehensible change to the original datapoint. 

\section{Towards communicative, contrastive explanations}
Such methods for computing contrastive explanations seek to provide contextually-relevant information to parties affected by a decision by describing how relevant closely related, alternative events could have occurred. However, choosing a relevant set of cases or events against which contrastive explanations are provided is not a straightforward challenge. The way in which information is transferred has a substantial impact on the quality and psychological acceptability of explanations \citep{Hilton1990Conversational}. The recipient's beliefs about an event to be explained are similarly constrained by the explainer's choice of explanation. As a result, the explainer's epistemological and normative values can have a significant effect on the recipient's understanding of an event.  \citet{Lombrozo2009Explanation} for example demonstrated that the type of explanation provided (e.g. mechanistic, functional) can influence the recipient's view of the importance of features in categorising a phenomenon \citep{poulin_visual_2006}. Explainers and explainees can have different motivations, for example to generate trust (explainer) or understand non-intuitive causes of a decision (explainee), meaning conflicts can arise in explanation selection and evaluation. 

The normativity of this relationship means that a risk exists of malicious explainers subtly discouraging explainees from critically questioning or contesting a decision through choice of explanation(s).  This risk is particularly acute when explanations are given to foster trust or understanding. A recipient's beliefs can potentially be `gamed' or manipulated to align with the explainer's preferred explanation of a phenomenon, meaning recipients can be `nudged' to take a preferred action (or not). Seemingly rational bases for otherwise unjustifiable decisions can for example be offered to nudge the recipient not to question or contest the decision. \citet{Lipton2016Mythos} has urged for caution in adopting post-hoc interpretation methods due to this potential to mislead recipients, for example by falsely attributing a decision to an irrelevant feature, or a more acceptable feature (e.g. post code, 'leadership') that serves as a proxy for a legally protected feature (cf. \citet{Barocas2016Big,kim_data-driven_2016}).

These risks may be mitigated by developing methods to provide contrastive, selective, and social explanations that not only enable information exchange and dialogue between giver and recipient, but critical argumentation and discussion of the justifiability of an event as well. This suggestion stems from prior work that suggests conversational explanations are essentially a form of argumentation.  Conversational explanations are used to both offer causes of an event and back claims as to the truth or relevance of these causes \citep{Antaki1992Explaining}. \citet{Walton2004new, Walton2007Dialogical} takes a similar approach in proposing a dialectical theory of everyday explanations, which suggests that giving explanations involves not only transfer of information or causal claims, but also argumentative support for these claims. 

These findings, that explanations are given via conversation and resemble argumentation, reveal a mechanistic link between explanation and justification as a type of discourse \citep{Fox2007Argumentation-based}. Justification is also often a discursive act, relying upon explanations to transfer knowledge and support claims made by each party. Interest in justification stems from an individual's ability to comprehend decisions made about them, and contest them when they are obviously incorrect or found unacceptable (or unjustifiable). As such, argumentation models of explanation resemble theories of justification and democracy based upon discourse, such as Jurgen Habermas' discourse ethics (located in his broader Theory of Communicative Action; \citep{Habermas1984Theory}).

Justification has recently received increasing attention by scholars discussing algorithmic accountability (e.g. \citet{Binns2017Algorithmic,biran_justification_2014,Weller2017Challenges,hildebrandt_challenges_2010,selbst_intuitive_2018}). However, justification currently occupies an uneasy position in xAI, in that it is rarely formally defined or related to explanations or ideals of transparency and accountability.\footnote{\citet{biran_justification_2014} prove an exception to this rule. They define explanation as an answer to the question "how did the system arrive at the prediction?" whereas justification answers the question "why should we believe the prediction is correct?" They further suggest that a satisfactory answer to the first question will also provide an answer to the second for explainees with sufficient expertise.} This unease is perhaps understandable; as \citet{Binns2017Algorithmic} notes, legitimate disagreements between epistemic and ethical standards for algorithmic decision-making can exist which require resolution, for example through democratic processes. Despite this, important conversations around the ethical acceptability or justifiability of algorithmic systems must occur in society if responsible deployment is sought. 

Finally, very little work in xAI addresses the link between interpretability and contestability of models or decisions \citep{Lipton2016Mythos}. Going forward, explanations of specific algorithmic decisions should allow the justification of a black-box model or decision to be debated and contested.. If we use justificatory discourse as a framework for explanation requirements, we need to determine what sort of records algorithmic systems must retain in order to allow for contesting and post-hoc auditing of abnormal events \citep{mittelstadt_automation_2016,sandvig_auditing_2014}, for instance through identification of classification errors or inaccurate input data \citep{poulin_visual_2006}. Where approximations are used to provide explanations, information should also be provided to affected parties detailing the limitations and relative resilience of the approximation, including the domain addressed and why it has been chosen. Further, meaningful, critical dialogue can be achieved between user, developer, and model by ensuring explanations are contrastive, selective, and social. We thus need to ensure xAI aims to develop methods for producing explanations of model functionality and specific decisions that embody these characteristics. Reliable methods for producing contrastive explanations and explanation by approximation are both required. 

At the moment, the xAI community largely fails in this task. Many approaches produce approximate and local models that are more akin to models in science. It may be possible to ask questions of these models, and to develop contrastive explanations from them, and so they are not without value. But they do not help us directly provide contrastive explanations to parties affected by algorithmic decisions. Going forward, the field must urgently close this gap.

\bibliographystyle{ACM-Reference-Format}
\bibliography{acmart}
\pagebreak

\end{document}